# DriVLM: Domain Adaptation of Vision-Language Models in Autonomous Driving

Xuran Zheng[O], Chang D. Yoo[*]

School of Electrical Engineering, KAIST

zxrshawn@kaist.ac.kr, cd_yoo@kaist.ac.kr

## Abstract

In recent years, large language models have had a very impressive performance, which largely contributed to the development and application of artificial intelligence, and the parameters and performance of the models are still growing rapidly. In particular, multimodal large language models (MLLM) can combine multiple modalities such as pictures, videos, sounds, texts, etc., and have great potential in various tasks. However, most MLLMs require very high computational resources, which is a major challenge for most researchers and developers. In this paper, we explored the utility of small-scale MLLMs and applied small-scale MLLMs to the field of autonomous driving. We hope that this will advance the application of MLLMs in real-world scenarios.

## 1. Introduction

Recent research has seen significant progress in multimodal large language models (MLLM) that combine large language models as well as visual models to achieve robust performance. These models are heavily trained on a variety of multimodal data and are able to efficiently align the latent space of the linguistic model with the visual representation of the visual model, thus showing strong performance in tasks such as visual comprehension and multiround dialog interaction. However, the lack of computational resources limits the application of MLLMs in real-world scenarios.

It is satisfying that some small-scale MLLMs are able to make a good balance between parameters and performance, get rid of the limitation of computational resources, and can be effectively applied in real-world scenarios. In order to connect the visual model with the LLMs, an intermediate layer is needed to align the visual model with the LLMs. This alignment may have the following drawbacks: (1) Different parameter scales. The LLM may reach more than 1000 billion parameters, while common VLLMs currently have only about 1 billion parameters. (2) Inefficient connection. The intermediate connection layer is usually limited in size and randomly initialized, and may not be able to capture cross-modal relationships in multimodal models. (3) Most existing MLLMs rely on vision encoders such as CLIP [1], which are trained on web-based image-text datasets and aligned with BERT [2]. Consequently, these encoders lack the ability to be aligned with the representations of LLMs. (4) Current approaches to adapting MLLMs to specialized domains primarily involve modifying model architecture, collecting large amounts of relevant training data, or specifying the training process to suit the specific domain. However, there is no unified framework for the downstream adaptation of LLMs, as different domains require distinct model configurations, data structures, and training procedures.

To address these challenges, it is essential to develop a powerful visual encoder with extensive visual knowledge and a generalized transfer learning scheme that can be efficiently applied to downstream tasks across various domains with low marginal costs. In this paper, we adopted InternVL as the baseline model, a series of small-scale yet highly capable MLLMs designed for easy adaptation to specialized domains. InternVL enhances small-scale visual encoder representations by initializing a 300M visual encoder with pre-trained weights from CLIP and employing InternViT-6B as a teacher model for knowledge distillation. Building upon this, the Mini-InternVL series with 1B, 2B, and 4B parameters was developed by integrating the visual encoder with pre-trained LLMs such as Qwen2-0.5B [3] and InternLM2-1.8B [4].

Equipped with a robust visual encoder, Mini-InternVL demonstrates strong performance on generalized multimodal benchmarks like MMBench [5] and ChartQA [6]. The proposed model could achieve great performance while possessing fewer parameters in the model, which significantly reduces the computational costs. To further optimize the model for domain-specific downstream tasks, they proposed a straightforward yet efficient transfer learning scheme. This scheme introduces a unified adaptation method applicable to a wide range of downstream tasks. We perform a comprehensive evaluation of our models through extensive experiments on both general and domain-specific benchmarks. The results demonstrate that this method enables the given model to achieve great performance on specific autonomous driving datasets.

---
[*] Corresponding Author

## 2. Related Works of Multimodal LLMs

### 2.1 Vision Large Language Models

Recent advancements in Vision Large Language Models (VLLMs) have enabled language models to process and interpret visual information. Models such as Flamingo [7], and MiniGPT-4 [8] introduced visual instruction tuning, while VisionLLM [9] incorporated visual grounding for tasks like localization. API-based approaches and embodied VLLMs like PaLM-E [10] and further expanded their applications.

Meanwhile, most vision foundation models (VFMs) used in MLLMs, such as CLIP [1], face limitations in scale and domain knowledge. Efforts like LLaVA-HR [11] adopt dual vision encoders to improve resolution but introduce complexity. InternViT addresses these issues with generative training across diverse datasets, achieving domain-general capabilities while reducing computational overhead. However, the overall progress of VFMs, critical to VLLMs, remains slower than desired.

### 2.2 Multimodal Large Language Models

With the rapid advancements in large language models (LLMs), multimodal large language models (MLLMs) have made significant progress. Early works framed multimodal understanding as a tool usage task, prompting LLMs to request captions from external models to process multimodal inputs. To better integrate pre-trained LLMs and vision foundation models (VFMs), some approaches introduced connectors to align their embedding spaces, achieving strong performance at a manageable cost. Other methods added extra layers to LLMs for vision feature fusion, reducing visual token input but increasing training overhead.

Recently, vision encoder-free architectures like MoMa [12] have utilized a single transformer to process visual and textual inputs simultaneously, simplifying deployment. However, many small-scale MLLMs, such as MiniCPM-V [13], are still limited to the natural image domain due to their reliance on CLIP-L [14] as the vision encoder, which is primarily trained on web image-text data and lacks domain diversity. In contrast, InternViT-300M is trained on a diverse range of image domains to enable robust generalization. Additionally, the success of open-source LLMs like LLaMA [15] has shown the potential of LLMs, integrating their capabilities with multi-modal interactions, and positioning MLLMs as a key research direction.

### 2.3 Domain Adaptation of MLLMs

The adaptation of large language models to specific domains has become a crucial area of development as industries seek to leverage AI models tailored to their needs. Domain specialization involves training or fine-tuning models using industry-specific data to improve their performance in specialized tasks, such as healthcare, legal, or finance sectors. This approach offers more accurate and efficient results compared to some general-purpose models, which may not capture the nuances or jargon of a specific field.

The challenge with domain specialization lies in ensuring LLMs effectively utilize their pre-trained knowledge while integrating domain-specific datasets. Techniques like transfer learning and domain adaptation are key methods used here. In transfer learning, a model pre-trained on general data is adapted to a target domain by fine-tuning its parameters with domain-specific data.

Recent efforts have demonstrated the benefits of this specialized approach. For instance, healthcare models like Claude are fine-tuned on medical data such as clinical notes and PubMed articles to assist professionals with tasks like summarizing patient histories and suggesting treatments. Similarly, the finance and legal sectors have seen models like ROSS' EVA, which is trained on legal documents to support professionals with tasks such as case law research and contract review. Various approaches have been explored to adapt Multimodal Large Language Models (MLLMs) to specific domains, such as GeoChat [16] for remote sensing, LLaVA-Med [17] for medical imaging, and DriveGPT4 [18] for autonomous driving. While these methods have shown promising outcomes, they typically require architectural modifications, large-scale domain-specific datasets, or custom training processes tailored to the target field. Despite these advancements, there is rarely a universally accepted framework for the downstream adaptation of MLLMs. Therefore, it is necessary to propose a simple yet effective transfer learning approach aimed at minimizing the inconsistencies among MLLMs across different domains, thus enhancing their interoperability.

Despite these challenges, domain-specialized LLMs hold the potential to revolutionize various industries by providing AI that deeply understands the language and complexities of specific domains, offering specific insights that are more accurate than those from general models.

## 3. Approaches and Experiment Results

### 3.1 Overall Architecture

Mini-InternVL is a small-scale multimodal model comprising three main components: a visual encoder, an MLP projector, and a large language model, as shown in Figure 1. The visual encoder is based on InternViT-300M, which has been optimized to support multiple versions of Mini-InternVL (e.g., Mini-InternVL-1B, 2B, and 4B) that integrate with different pre-trained language models. The model employs a dynamic resolution input strategy to improve its ability to capture fine-grained details and uses a pixel fixing operation to significantly reduce the number of visual tokens, enabling efficient processing of high-resolution images.

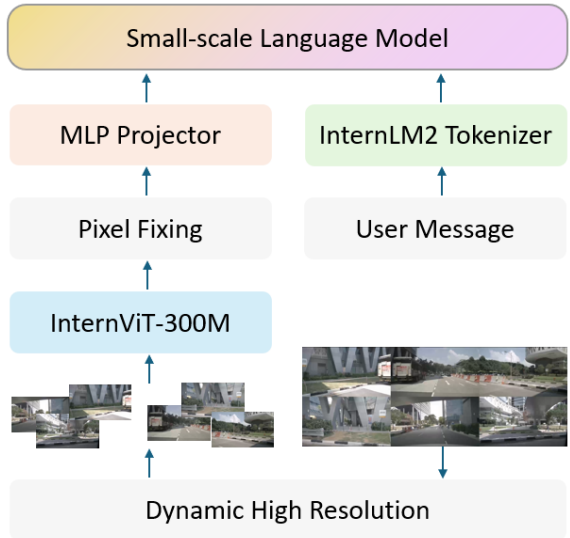

Figure 1. General architecture of Mini-InternVL

The training process consists of two stages. In the first stage, the visual encoder remains fixed while only the MLP is adjusted to align image and language representations. This stage leverages large and diverse datasets covering multiple tasks for robust pre-training. In the second stage, full-parameter fine-tuning is conducted to enhance the model's performance on multimodal tasks, including image captioning, text recognition, chart interpretation, and cross-domain reasoning, further enriching its knowledge and instruction-following capabilities.

## 3.2 Alignment Strategy and Domain Adaption

Although significant progress has been made in applying Multimodal Large Language Models (MLLMs) to downstream tasks, the absence of a universal adaptation framework remains a key challenge. Differences in model designs, data formats, and training strategies across domains lead to significant differences among MLLMs, complicating their standardization. To address this problem, Mini-InternVL is a simple yet effective transfer learning framework.

Instruction fine-tuning plays a critical role in teaching models to understand user instructions, typically converting data into Visual Question Answering (VQA) or dialogue formats. For traditional tasks, we reformat them into VQA format using the following methods:

**Image Classification Tasks**: Classification tasks are reformulated as multiple-choice questions. For instance, given a set of candidate labels and a ground truth label, prompts like "Classify the image into one of the following categories: ..." guide the model to perform the task.

**Visual Localization Tasks**: Objects are marked using <ref></ref> tags for their names and <box></box> for their coordinates, such as [[x1, y1, x2, y2]]. These tags enable the model to produce precise localization outputs.

**Region-Aware Tasks**: These tasks involve providing spatial information to the model, requiring it to focus on a specific region. The target area is represented using <box>[[x1, y1, x2, y2]]</box>, guiding the model to generate descriptions or answers related to the specified region.

**Multi-View Images**: For autonomous driving, we process multi-view images by dynamically adjusting their resolution and combining them into a unified format, supplemented with camera position annotations.

During domain adaptation, it is necessary to fine-tune Mini-InternVL with all parameters, integrating a balanced proportion of general multimodal data. This approach retains domain-specific performance while improving generalization across tasks. By adjusting the ratio of general data to domain-specific data, Mini-InternVL achieves an optimal balance between performance and computational efficiency.

## 3.3 Experiment Results

In this experiment, Mini-InternVL is pre-trained on DriveLM-nuScenes 1.1, which includes 317K samples covering perception, prediction, and planning tasks, providing a comprehensive representation of autonomous driving scenarios. Images are captured from six aspects and processed using dynamic resolution by dividing them into 448×448 pixels, while a thumbnail is also generated to provide a global view. Additionally, each viewpoint image is labeled with text indicating its camera position, such as "CAM_FRONT."

We tested our model using the dataset DriveLM-nuScenes-version-1.1-val, and the results are shown in Table 1. The CIDEr (Consensus-based Image Description Evaluation) score of the Mini-InternVL2B model is 0.191, which can be considered an excellent result of domain adaptation. The model performed slightly lower on the Bleu 1 metric, which might be due to its limitations in predicting object center points. Furthermore, the constraints of the training data and evaluation metrics could impact the potential of larger-scale models.

Table 1. Test results of main models for autonomous driving on the DriveLM dataset

| Model | Acc. | Bleu 1 | CIDEr |
|---|---|---|---|
| MTMM | 0.747 | 0.76 | 0.18 |
| Team NVIDIA | 0.775 | - | - |
| MMFM_AD | 0.666 | - | - |
| Mini-InternVL-DA-1B | 0.701 | 0.736 | 0.167 |
| Mini-InternVL-DA-2B | 0.763 | 0.762 | 0.191 |

## 4. Conclusion

In this paper, we introduce Mini-InternVL, a collection of small-scale and open-source Multimodal Large Language Models (MLLMs) designed for efficient deployment in environments with limited resources. Mini-InternVL employs InternViT-300M as a compact vision encoder and leverages knowledge distillation from more advanced teacher models to incorporate knowledge from diverse domains, addressing the constraints of encoders like CLIP-ViT. With substantially fewer parameters, Mini-InternVL achieves approximately 90% of the performance of larger models and demonstrates strong capabilities in tasks of other fields such as domain-specific visual understanding. To enhance the application of small-scale multimodal models in specialized applications, a unified transfer framework is adopted, enabling seamless adaptation to various domains while achieving competitive results compared to other specific approaches. We believe this work offers meaningful perspectives on the application and development of MLLMs.

## Acknowledgments

This work was partly supported by Institute for Information & communications Technology Planning & Evaluation (IITP) grant funded by the Korea government (MSIT) (No.RS-2021-II211381, Development of Causal AI through Video Understanding and Reinforcement Learning, and Its Applications to Real Environments) and partly supported by the National Research Foundation of Korea (NRF) grant funded by the Korea government (MSIT) (No. 2022R1A2C2012706).

## References

[1] Alec Radford, Jong Wook Kim, Chris Hallacy, Aditya Ramesh, Gabriel Goh, Sandhini Agarwal, Girish Sastry, Amanda Askell, Pamela Mishkin, and Jack Clark. Learning transferable visual models from natural language supervision. In ICML, 2021.

[2] Jacob Devlin. Bert: Pre-training of deep bidirectional transformers for language understanding. arXiv preprint arXiv:1810.04805, 2018.

[3] An Yang, Baosong Yang, Binyuan Hui, Bo Zheng, Bowen Yu, Chang Zhou, Chengpeng Li, Chengyuan Li, Dayiheng Liu, Fei Huang, et al. Qwen 2 technical report. arXiv preprint arXiv:2407.10671, 2024.

[4] Zheng Cai, Maosong Cao, Haojiong Chen, Kai Chen, Keyu Chen, Xin Chen, Xun Chen, Zehui Chen, Zhi Chen, Pei Chu, et al. Internlm2 technical report. arXiv preprint arXiv:2403.17297, 2024.

[5] Yuan Liu, Haodong Duan, Yuanhan Zhang, Bo Li, Songyang Zhang, Wangbo Zhao, Yike Yuan, Jiaqi Wang, Conghui He, Ziwei Liu, et al. Mmbench: Is your multi-modal model an all-around player? arXiv preprint arXiv:2307.06281, 2023.

[6] Ahmed Masry, Xuan Long Do, Jia Qing Tan, Shafiq Joty, and Enamul Hoque. Chartqa: A benchmark for question answering about charts with visual and logical reasoning. In ACL, pages 2263–2279, 2022.

[7] Jean-Baptiste Alayrac, Jeff Donahue, Pauline Luc, Antoine Miech, Iain Barr, Yana Hasson, Karel Lenc, Arthur Mensch, Katherine Millican, Malcolm Reynolds, et al. Flamingo: a visual language model for few-shot learning. NeurIPS, 35:23716–23736, 2022. 1, 3, 8

[8] Deyao Zhu, Jun Chen, Xiaoqian Shen, Xiang Li, and Mohamed Elhoseiny. Minigpt-4: Enhancing vision-language understanding with advanced large language models. arXiv preprint arXiv:2304.10592, 2023. 1, 3, 11

[9] Wenhai Wang, Zhe Chen, Xiaokang Chen, Jiannan Wu, Xizhou Zhu, Gang Zeng, Ping Luo, Tong Lu, Jie Zhou, Yu Qiao, et al. Visionllm: Large language model is also an open-ended decoder for vision-centric tasks. NeurIPS, 2023. 1, 3

[10] Danny Driess, Fei Xia, Mehdi SM Sajjadi, Corey Lynch, Aakanksha Chowdhery, Brian Ichter, Ayzaan Wahid, Jonathan Tompson, Quan Vuong, Tianhe Yu, et al. Palm-e: An embodied multimodal language model. arXiv preprint arXiv:2303.03378, 2023. 3

[11] Xiaohua Zhai, Basil Mustafa, Alexander Kolesnikov, and Lucas Beyer. Sigmoid loss for language image pre-training. In ICCV, pages 11975–11986, 2023.

[12] Xi Victoria Lin, Akshat Shrivastava, Liang Luo, Srinivasan Iyer, Mike Lewis, Gargi Ghosh, Luke Zettlemoyer, and Armen Aghajanyan. Moma: Efficient early-fusion pre-training with mixture of modality-aware experts. arXiv preprint arXiv:2407.21770, 2024.

[13] Yuan Yao, Tianyu Yu, Ao Zhang, Chongyi Wang, Junbo Cui, Hongji Zhu, Tianchi Cai, Haoyu Li, Weilin Zhao, Zhihui He, et al. Minicpm-v: A gpt-4v level mllm on your phone. arXiv preprint arXiv:2408.01800, 2024.

[14] Alec Radford, Jong Wook Kim, Chris Hallacy, Aditya Ramesh, Gabriel Goh, Sandhini Agarwal, Girish Sastry, Amanda Askell, Pamela Mishkin, and Jack Clark. Learning transferable visual models from natural language supervision. In ICML, 2021.

[15] Hugo Touvron, Thibaut Lavril, Gautier Izacard, Xavier Martinet, Marie-Anne Lachaux, Timothée Lacroix, Baptiste Rozière, Naman Goyal, Eric Hambro, Faisal Azhar, et al. Llama: Open and efficient foundation language models. arXiv preprint arXiv:2302.13971, 2023. 2, 3

[16] Kartik Kuckreja, Muhammad S. Danish, Muzammal Naseer, Abhijit Das, Salman Khan, and Fahad S. Khan. Geochat: Grounded large vision-language model for remote sensing. CVPR, 2024.

[17] Chunyuan Li, Cliff Wong, Sheng Zhang, Naoto Usuyama, Haotian Liu, Jianwei Yang, Tristan Naumann, Hoifung Poon, and Jianfeng Gao. Llavamed: Training a large language-and-vision assistant for biomedicine in one day. arXiv preprint arXiv:2306.00890, 2023.

[18] Zhenhua Xu, Yujia Zhang, Enze Xie, Zhen Zhao, Yong Guo, Kwan-Yee K. Wong, Zhenguo Li, and Hengshuang Zhao. Drivegpt4: Interpretable end-to-end autonomous driving via large language model. arXiv preprint arXiv:2310.01412, 2023.